\documentclass{article} % For LaTeX2e
\usepackage[accepted]{icml2020}
\usepackage{mathtools}
\usepackage{url}
\usepackage[paperVersion]{marcpaper}

\author{Marc Brockschmidt\\
Microsoft Research\\
Cambridge, United Kingdom\\
\texttt{mabrocks@microsoft.com} 
}

\begin{document}

\twocolumn[
\icmltitle{GNN-FiLM: Graph Neural Networks with Feature-wise Linear Modulation}
\begin{icmlauthorlist}
  \icmlauthor{Marc Brockschmidt}{msrc}
\end{icmlauthorlist}
\icmlaffiliation{msrc}{Microsoft Research, Cambridge, UK}
\icmlcorrespondingauthor{Marc Brockschmidt}{mabrocks@microsoft.com}
\icmlkeywords{Graph Neural Networks}
\vskip 0.3in
]

\printAffiliationsAndNotice{}

\begin{abstract}
 This paper presents a new Graph Neural Network (GNN) type using feature-wise
 linear modulation (FiLM).
 Many standard GNN variants propagate information along the edges of a graph by
 computing messages based only on the representation of the source of
 each edge.
 In GNN-FiLM, the representation of the target node of an edge is
 used to compute a transformation that can be applied to all incoming
 messages, allowing feature-wise modulation of the passed information.
 Different GNN architectures are compared in extensive experiments on
 three tasks from the literature, using re-implementations of many
 baseline methods.
 Hyperparameters for all methods were found using extensive search,
 yielding somewhat surprising results: differences between
 state of the art models are much smaller than reported in the
 literature and well-known simple baselines that are often not compared to
 perform better than recently proposed GNN variants.
 Nonetheless, GNN-FiLM outperforms these methods on a regression task
 on molecular graphs and performs competitively on other tasks.
\end{abstract}

\section{Introduction}
\label{sect:introduction}
Learning from graph-structured data has seen explosive growth over the
last few years, as graphs are a convenient formalism to model the broad
class of data that has objects (treated as vertices) with some known
relationships (treated as edges).
This capability has been used in
 physical and biological systems,
 knowledge bases,
 computer programs,
 and relational reasoning in computer vision tasks.
This graph construction is a highly complex form of feature engineering,
mapping the knowledge of a domain expert into a graph structure which can
be consumed and exploited by high-capacity neural network
models.

Most neural graph learning methods can be summarised as neural message
passing~\citep{gilmer2017neural}: nodes are initialised with some
representation and then exchange information by transforming their current
state (in practice with a single linear layer) and sending it as a message to
all neighbours in the graph.
At each node, messages are aggregated in some way and then used to update
the associated node representation.
In this setting, the message is entirely determined by the source node
(and potentially the edge type) and the target node is not taken into
consideration.
A (partial) exception to this is the family of Graph Attention
Networks~\citep{velickovic2018graph}, where the agreement between
source and target representation of an edge is used to determine the
\emph{weight} of the message in an attention architecture.
However, this weight is applied to all dimensions of the message at
the same time.

A simple consequence of this observation may be to simply compute messages
from the pair of source and target node state.
However, the linear layer commonly used to compute messages would only
allow additive interactions between the representations of source and target
nodes.
More complex transformation functions are often impractical, as computation
in GNN implementations is dominated by the message transformation function.

However, this need for non-trivial interaction between different information
sources is a common problem in neural network design.
A recent trend has been the use of \emph{hypernetworks}~\citep{ha2016HyperNetworks},
neural networks that compute the weights of other networks.
In this setting, interaction between two signal sources is achieved by
using one of them as the input to a hypernetwork and the other as input
to the computed network.
While an intellectually pleasing approach, it is often impractical because
the prediction of weights of non-trivial neural networks is computationally
expensive.
Approaches to mitigate this exist (e.g., \citet{wu2018pay} handle this in
natural language processing), but are often domain-specific.

A more general mitigation method is to restrict the structure of the computed
network.
Recently, ``feature-wise linear modulations'' (FiLM) were introduced in the
visual question answering domain~\citep{perez2017film}.
Here, the hypernetwork is fed with an encoding of a question and produces
an element-wise affine function that is applied to the features extracted
from a picture.
This can be adapted to the graph message passing domain by using the
representation of the target node to compute the affine function.
This compromise between expressiveness and computational feasibility has been
very effective in some domains and the results presented in this article
indicate that it is also a good fit for the graph domain.

This article explores the use of hypernetworks in learning on graphs.
\rSC{sect:model} first reviews existing GNN models from the related work
to identify commonalities and differences.
This involves generalising a number of existing formalisms to new
formulations that are able to handle graphs with different types of edges,
which are often used to model different relationship between vertices.
Then, two new formalisms are introduced:
\underline{R}elational \underline{G}raph \underline{D}ynamic \underline{C}onvolutional
\underline{N}etworks (RGDCN), which dynamically compute the neural message
passing function as a linear layer, and
\underline{G}raph \underline{N}eural \underline{N}etworks with
\underline{F}eature-w\underline{i}se \underline{L}inear \underline{M}odulation (GNN-FiLM),
which combine learned message passing functions with dynamically computed
element-wise affine transformations.
In \rSC{sect:evaluation}, a range of baselines are compared in extensive 
experiments on three tasks from the literature, spanning classification,
regression and ranking tasks on small and large graphs.

The core contributions of this article are 
 (1) a new GNN type based on the FiLM idea from visual question answering,
 (2) generalisations of existing GNN types (GAT and GIN) to the multi-relational setting,
and
 (3) an empirical evaluation using a unified framework and a consistent hyperparameter optimisation method.
The key takeaways are that
 (a) existing GNN models perform similarly on many tasks,
 (b) a simple GNN-MLP baseline model, using MLPs on the concatenation of source and target representation, often outperforms well-published models from the literature,
 (c) GNN-FiLM is competitive with or outperforms GNN-MLP on all tested tasks.

\section{Model}
\label{sect:model}
\paragraph{Notation.}
Let $\edgetypes$ be a finite (usually small) set of edge types.
Then, a directed graph $\graph = (\nodes, \edges)$ has nodes $\nodes$ and typed edges
$\edges \subseteq \nodes \times \edgetypes \times \nodes$, where
$(u, \ell, v) \in \edges$ denotes an edge from node $u$ to node $v$ of type $\ell$,
usually written as $\typededge{u}{\ell}{v}$.
For example, in a graph representation of source code, successive tokens may be
connected using a \code{NextToken} edge type, and method calls may be connected
to the corresponding method definition using \code{CalledMethod} edges.

\colorlet{nodeAcol}{orange!80!red}
\colorlet{nodeBcol}{blue!50}
\colorlet{nodeCcol}{violet!50}
\colorlet{nodeDcol}{green}
\colorlet{edge0col}{ForestGreen}
\colorlet{edge1col}{SeaGreen!50!brown}
\colorlet{edge2col}{purple}

\begin{figure}
  \resizebox{\columnwidth}{!}{
    \begin{tikzpicture}[gnnex]
      \node[vertex,fill=nodeAcol] (A) at (0,0) {$A$};
      \node[vertex,fill=nodeBcol]   (B) at ($(A) + (1,1)$) {$B$};
      \node[vertex,fill=nodeCcol] (C) at ($(A) + (1,-1)$) {$C$};
      \node[vertex,fill=nodeDcol]  (D) at ($(A) + (2,0)$) {$D$};
      \draw
       (B) edge[->, edge1col] node[above] {\scriptsize 1} (A)
       (B) edge[->, edge1col] node[right, yshift=3mm, xshift=-1mm] {\scriptsize 1} (C)
       (D) edge[->, edge1col] node[below, yshift=1mm, xshift=2mm] {\scriptsize 1} (A)
       (C) edge[->, edge1col] node[below] {\scriptsize 1} (D)
       (B) edge[->, edge2col] node[above] {\scriptsize 2} (D)
       (C) edge[->, edge2col] node[below] {\scriptsize 2} (A)
       (A) edge[->, loop left, looseness=4, edge0col] node[left] {\scriptsize $\circlearrowleft$} (A)
       (B) edge[->, loop right, looseness=4, edge0col] node[right] {\scriptsize $\circlearrowleft$} (B)
       (C) edge[->, loop right, looseness=4, edge0col] node[right] {\scriptsize $\circlearrowleft$} (C)
       (D) edge[->, loop right, looseness=4, edge0col] node[right] {\scriptsize $\circlearrowleft$} (D)
      ;

      \node[vertex,fill=nodeAcol] (AP) at ($(A) + (4.5, 0)$) {$A'$};
      \node[vertex,black!40,fill=nodeBcol!40]   (BP) at ($(AP) + (1,1)$) {$B'$};
      \node[vertex,black!40,fill=nodeCcol!40] (CP) at ($(AP) + (1,-1)$) {$C'$};
      \node[vertex,black!40,fill=nodeDcol!40]  (DP) at ($(AP) + (2,0)$) {$D'$};
      \draw
       (BP) edge[->, edge1col!20] node[above] {\scriptsize 1} (AP)
       (BP) edge[->, edge1col!20] node[right, yshift=3mm, xshift=-1mm] {\scriptsize 1} (CP)
       (DP) edge[->, edge1col!20] node[below, yshift=1mm, xshift=2mm] {\scriptsize 1} (AP)
       (CP) edge[->, edge1col!20] node[below] {\scriptsize 1} (DP)
       (BP) edge[->, edge2col!20] node[above] {\scriptsize 2} (DP)
       (CP) edge[->, edge2col!20] node[below] {\scriptsize 2} (AP)     
       (AP) edge[->, loop left, looseness=4, edge0col!20] node[left] {\scriptsize $\circlearrowleft$} (AP)
       (BP) edge[->, loop right, looseness=4, edge0col!20] node[right] {\scriptsize $\circlearrowleft$} (BP)
       (CP) edge[->, loop right, looseness=4, edge0col!20] node[right] {\scriptsize $\circlearrowleft$} (CP)
       (DP) edge[->, loop right, looseness=4, edge0col!20] node[right] {\scriptsize $\circlearrowleft$} (DP)  
      ;

      \draw
       ($(A.north)$) edge[->, dashed, edge0col, bend left=80] node[below, xshift=4ex] {\scriptsize $\circlearrowleft$} ($(AP.north)$)
       ($(B.north east)$) edge[->, dashed, edge1col, bend left=20] node[below] {\scriptsize 1} ($(AP.north west)$)
       ($(D.north east)$) edge[->, dashed, edge1col, bend left=20] node[below] {\scriptsize 1} ($(AP.north west)$)
       ($(C.south east)$) edge[->, dashed, edge2col, bend right=20] node[above] {\scriptsize 2} ($(AP.south west)$)
      ;
    \end{tikzpicture}
  }
  \caption{Graphical illustration of GNN computation.
    Left: Graph of vertices 
    $\textcolor{nodeAcol!90!black}{A},
      \textcolor{nodeBcol!80!black}{B},
      \textcolor{nodeCcol!80!black}{C},
      \textcolor{nodeDcol!80!black}{D}$
    (identified with their representation) and directed edge types
    $\textcolor{edge0col!90!black}{\circlearrowleft},
      \textcolor{edge1col!90!black}{1},
      \textcolor{edge2col!90!black}{2}$.
    Right: Graph with same topology, but new representations. Data flow in computation of $\textcolor{nodeAcol!90!black}{A'}$ indicated by dashed edges.
    Best viewed in colour.
    \label{fig:gnn-comp-sketch}
  }
\end{figure}

\paragraph{Graph Neural Networks.}
As discussed above, Graph Neural Networks operate by propagating information
along the edges of a given graph.
Concretely, each node $\node$ is associated with an initial representation
$\state{\node}{0}$ (for example obtained from the label of that node, or by
some other model component).
Then, a GNN layer updates the node representations using the node representations 
of its neighbours in the graph, yielding representations $\state{\node}{1}$.
This computation is illustrated for an example graph in \rF{fig:gnn-comp-sketch}.
The process can be unrolled through time by repeatedly applying the same
update function, yielding representations $\state{\node}{2} \ldots \state{\node}{T}$.
Alternatively, several GNN layers can be stacked, which is intuitively similar to
unrolling through time, but increases the
GNN capacity by using different parameters for each timestep.

In Gated Graph Neural Networks (GGNN)~\citep{li2015gated}, the update rule uses
one linear layer $\weights_{\!\ell}$ per edge type $\ell$ to compute messages and
combines the aggregated messages with the current representation of a node using
a recurrent unit $r$ (e.g., GRU or LSTM cells), yielding the following definition.
\begin{align}
  \state{v}{t+1} & = r\left(\state{v}{t},
                            \sum_{\typededge{u}{\ell}{v} \in \edges} 
                               \weights_{\!\ell} \state{u}{t}
                            \; ; \; \parameters_r
                      \right) \label{eq:ggnn}
\end{align}
The learnable parameters of the model are the edge-type-dependent weights
$\weights_{\!\ell}$ and the recurrent cell parameters $\parameters_r$.

In Relational Graph Convolutional Networks (R-GCN)~\citep{schlichtkrull2017modeling}
(an extension of Graph Convolutional Networks (GCN)~\citep{kipf2017semi}),
the gated unit is replaced by a simple non-linearity $\sigma$ (e.g., the hyperbolic
tangent).
\begin{align}
\state{v}{t+1} & = \sigma\left(\sum_{\typededge{u}{\ell}{v} \in \edges} 
                                  \frac{1}{c_{v,\ell}} \cdot \weights_{\!\ell} \state{u}{t}
                         \right) \label{eq:rgcn}
\end{align}
Here, $c_{v,\ell}$ is a normalisation factor usually set to the number of edges of
type $\ell$ ending in $v$.
The learnable parameters of the model are the edge-type-dependent weights
$\weights_{\!\ell}$.
It is important to note that in this setting, the edge type set $\edgetypes$
is assumed to contain a special edge type $\circlearrowleft$ for self-loops $\typededge{v}{\circlearrowleft}{v}$,
allowing state associated with a node to be kept.

In Graph Attention Networks (GAT)~\citep{velickovic2018graph}, new node
representations are computed from a weighted sum of neighbouring node
representations.
The model can be generalised from the original definition to support
different edge types as follows (we will call this R-GAT below).\footnote{Note 
 that this is similar to the ARGAT model presented by \citet{busbridge19rgat},
 but unlike the models studied there (and like the original GATs)
 uses a single linear layer to compute attention scores $e_{u, \ell, v}$,
 instead of simpler additive or multiplicative variants.}
\begin{align}
e_{u, \ell, v}     & = \text{LeakyReLU}(\boldsymbol\alpha_\ell \cdot (\weights_{\!\ell} \state{u}{t} \Vert \weights_{\!\ell} \state{v}{t})) \nonumber\\
\boldsymbol a_v    & = \text{softmax}(e_{u, \ell, v} \mid \typededge{u}{\ell}{v} \in \edges) \nonumber\\
\state{v}{t+1}     & = \sigma\left(\sum_{\typededge{u}{\ell}{v} \in \edges} 
                                     (\boldsymbol a_v)_{\typededge{u}{\ell}{v}} \cdot \weights_{\!\ell} \state{u}{t}
                              \right) \label{eq:gat}
\end{align}
Here, $\boldsymbol\alpha_\ell$ is a learnable row vector used to weigh different feature
dimensions in the computation of an attention (``relevance'') score of the node
representations,
$\boldsymbol x \Vert \boldsymbol y$ is the concatenation of vectors $\boldsymbol x$ and $\boldsymbol y$, and 
$(\boldsymbol a_v)_{\typededge{u}{\ell}{v}}$ refers to the weight computed by the softmax
for that edge.
The learnable parameters of the model are the edge-type-dependent weights
$\weights_{\!\ell}$ and the attention parameters $\boldsymbol\alpha_\ell$.
In practice, GATs usually employ several attention heads that independently
implement the mechanism above in parallel, using separate learnable
parameters.
The results of the different attention heads are then concatenated after each
propagation round to yield the value of $\state{v}{t+1}$.

More recently, \citet{xu2019how} analysed the expressiveness of different GNN
types, comparing their ability to distinguish similar graphs with the
Weisfeiler-Lehman (WL) graph isomorphism test.
Their results show that GCNs and the GraphSAGE model~\citep{hamilton2017inductive}
are strictly weaker than the WL test and hence they developed Graph Isomorphism
Networks (GIN), which are indeed as powerful as the WL test.
While the GIN definition is limited to a single edge type, Corollary 6 of
\citet{xu2019how} shows that using the definition
\begin{align*}
  \state{v}{t+1} & = \varphi\left((1 + \epsilon) \cdot f(\state{v}{t})
                                  + \sum_{\typededge{u}{}{v} \in \edges} f(\state{u}{t})
                            \right) \, ,
\end{align*}
there are choices for $\epsilon$, $\varphi$ and $f$ such that the node
representation update is sufficient for the overall network to be as 
powerful as the WL test.
In the setting of different edge types, the function $f$ in the sum over neighbouring
nodes needs to reflect different edge types to distinguish graphs such as
 $v \tightlabelledarrow{1} u \tightlabelledleftarrow{2} w$
and
 $v \tightlabelledarrow{2} u \tightlabelledleftarrow{1} w$
from each other.
Using different functions $f_\ell$ for different edge types makes it possible
to unify the use of the current node representation $\state{v}{t}$ with the use of
neighbouring node representations by again using a fresh edge type $\circlearrowleft$ for
self-loops $\typededge{v}{\circlearrowleft}{v}$.
In that setting, the factor $(1 + \epsilon)$ can be integrated into $f_\circlearrowleft$.
Finally, following an argument similar to \citet{xu2019how}, $\varphi$ and
$f$ at subsequent layers can be ``merged'' into a single function which can
be approximated by a multilayer perceptron (MLP), yielding the final R-GIN
definition
\begin{align}
  \state{v}{t+1} & = \sigma\left(\sum_{\typededge{u}{\ell}{v} \in \edges}
                                 \mathit{MLP}_{\ell}(\state{u}{t})
                           \right). \label{eq:rgin}
\end{align}
The learnable parameters here are the edge-specific MLPs $\mathit{MLP}_{\ell}$.
Note that Eq. \rEq{eq:rgin} is very similar to the definition of R-GCNs (Eq. \rEq{eq:rgcn}),
only dropping the normalisation factor $\frac{1}{c_{v,\ell}}$ and replacing linear
layers by an MLP.

While many more GNN variants exist, the four formalisms above are broadly
representative of general trends.
It is notable that in all of these models, the information passed from one
node to another is based on the learned weights and the representation
of the source of an edge.
In contrast, the representation of the \emph{target} of an edge is only 
updated (in the GGNN case Eq. \rEq{eq:ggnn}), treated as another incoming message
(in the R-GCN case Eq. \rEq{eq:rgcn} and the R-GIN case Eq. \rEq{eq:rgin}),
or used to weight the relevance of an edge (in the R-GAT case Eq. \rEq{eq:gat}).
Sometimes unnamed GNN variants of the above are used (e.g., by 
\citet{selsam19learning,paliwal19graph}), replacing the linear
layers to compute the messages for each edge by MLPs applied to the
concatenation of the representations of source and target nodes.
In the experiments, this will be called GNN-MLP, formally defined as follows.\footnote{
  These are similar to R-GIN, but apply an MLP to the concatenation of source and target
  state for each message.}
\begin{align}
  \state{v}{t+1} & = \sigma\left(\sum_{\typededge{u}{\ell}{v} \in \edges} 
                                    \frac{1}{c_{v,\ell}}
                                      \cdot
                                        \mathit{MLP}_{\ell}\left(
                                                          \state{u}{t} \Vert \state{v}{t}
                                                    \right)
                           \right) \label{eq:gnn_edge_mlp}
\end{align}
Below, we will instantiate the $\mathit{MLP}_\ell$ with a single linear layer
to obtain what we call GNN-MLP0, which only differs from R-GCNs (Eq. \rEq{eq:rgcn})
in that the message passing function is applied to the concatenation of
source and target state.

\subsection{Graph Hypernetworks}
Hypernetworks (i.e., neural networks computing the parameters of another
neural network)~\citep{ha2016HyperNetworks} have been successfully applied to
a number of different tasks; naturally raising the question if they are also
applicable in the graph domain.

Intuitively, a hypernetwork corresponds to a higher-order function, i.e.,
it can be viewed as a function computing another function.
Hence, a natural idea would be to use the target of a message propagation
step to compute the function computing the message; essentially
allowing it to focus on features that are especially relevant for the
update of the target node representation.

\paragraph{\underline{R}elational \underline{G}raph \underline{D}ynamic \underline{C}onvolutional \underline{N}etworks (RGDCN)}
A first attempt would be to adapt \rEq{eq:rgcn} to replace the learnable
message transformation $\weights_{\!\ell}$ by the result of some learnable
function $f$ that operates on the target representation:
\begin{align*}
  \state{v}{t+1} & = \sigma\left(\sum_{\typededge{u}{\ell}{v} \in \edges} 
                                    f(\state{v}{t} \; ; \; \parameters_{f, \ell}) \state{u}{t}
                            \right)
\end{align*}
However, for a representation size $D$, $f$ would need to produce a matrix
of size $D^2$ from $D$ inputs.
Hence, if implemented as a simple linear layer, $f$ would have on the order
of $\mathcal{O}(D^3)$ parameters, quickly making it impractical in most contexts.

Following \citet{wu2018pay}, this can be somewhat mitigated by splitting the node
representations \state{v}{t} into $C$ ``chunks'' \state{v, c}{t} of dimension
$K = \frac{D}{C}$:
\begin{align}
  \weights_{\!\ell, t, v, c}
                 & = f(\state{v}{t} \; ; \; \parameters_{f, \ell, c})\nonumber\\
  \state{v}{t+1} & = 
    \mathop{\big\Vert}_{1 \leq c \leq C}
      \sigma\left(
        \sum_{\typededge{u}{\ell}{v} \in \edges}
        \weights_{\!\ell, t, v, c} \state{u, c}{t}
      \right) \label{eq:rgdcn}
\end{align}
The number of parameters of the model can now be reduced by tying the value
of some instances of $\parameters_{f, \ell, c}$.
For example, the update function for a chunk $c$ can be computed using
only the corresponding chunk of the node representation \state{v, c}{t}, or
the same update function can be applied to all ``chunks'' by setting
$\parameters_{f, \ell, 1} = \ldots = \parameters_{f, \ell, C}$.
The learnable parameters of the model are only the hypernetwork parameters 
$\parameters_{f, \ell, c}$.
This is somewhat less desirable than the related idea of \citet{wu2018pay},
which operates on sequences, where sharing between neighbouring elements of
the sequence has an intuitive interpretation that is not applicable in the
general graph setting.

\begin{figure*}[t]
  \resizebox{1\textwidth}{!}{
  \begin{minipage}{1\textwidth}
   \begin{alignat*}{11}
    \text{GGNN: }
      & \textcolor{nodeAcol!90!black}{A'} = 
        &GRU&(
        &                                                             \textcolor{nodeAcol!90!black}{A}\phantom{)}
        &\ , \ & \weights_{\textcolor{edge1col!90!black}{\!1}}           \cdot \textcolor{nodeBcol!90!black}{B}\phantom{)}
        &+& \weights_{\textcolor{edge2col!90!black}{\!2}}                \cdot \textcolor{nodeCcol!90!black}{C}\phantom{)}
        &+& \weights_{\textcolor{edge1col!90!black}{\!1}}                \cdot \textcolor{nodeDcol!90!black}{D}\phantom{)}
        &)\\    
    \text{R-GCN: }
      & \textcolor{nodeAcol!90!black}{A'} = 
        &\sigma&(
        &   \weights_{\textcolor{edge0col!90!black}{\!\circlearrowleft}} \cdot \textcolor{nodeAcol!90!black}{A}\phantom{)}
        &+& \weights_{\textcolor{edge1col!90!black}{\!1}}                \cdot \textcolor{nodeBcol!90!black}{B}\phantom{)}
        &+& \weights_{\textcolor{edge2col!90!black}{\!2}}                \cdot \textcolor{nodeCcol!90!black}{C}\phantom{)}
        &+& \weights_{\textcolor{edge1col!90!black}{\!1}}                \cdot \textcolor{nodeDcol!90!black}{D}\phantom{)}
        &)\\
    \text{R-GAT: }
      & \textcolor{nodeAcol!90!black}{A'} = 
        &\sigma&(
        &   (\boldsymbol a_{\textcolor{nodeAcol!90!black}{A'}})_{
              \typededge{\textcolor{nodeAcol!90!black}{A}}%
                        {\textcolor{edge0col!90!black}{\circlearrowleft}}%
                        {\textcolor{nodeAcol!90!black}{A}}}
              \cdot
             \weights_{\textcolor{edge0col!90!black}{\!\circlearrowleft}}
              \cdot
             \textcolor{nodeAcol!90!black}{A}\phantom{)}
        &+& (\boldsymbol a_{\textcolor{nodeAcol!90!black}{A'}})_{
              \typededge{\textcolor{nodeBcol!90!black}{B}}%
                        {\textcolor{edge1col!90!black}{1}}%
                        {\textcolor{nodeAcol!90!black}{A}}}
              \cdot
             \weights_{\textcolor{edge1col!90!black}{\!1}} 
              \cdot \textcolor{nodeBcol!90!black}{B}\phantom{)}
        &+& (\boldsymbol a_{\textcolor{nodeAcol!90!black}{A'}})_{
              \typededge{\textcolor{nodeCcol!90!black}{C}}%
                        {\textcolor{edge2col!90!black}{2}}%
                        {\textcolor{nodeAcol!90!black}{A}}}
              \cdot
             \weights_{\textcolor{edge2col!90!black}{\!2}}
              \cdot
             \textcolor{nodeCcol!90!black}{C}\phantom{)}
        &+& (\boldsymbol a_{\textcolor{nodeAcol!90!black}{A'}})_{
              \typededge{\textcolor{nodeDcol!90!black}{D}}%
                        {\textcolor{edge1col!90!black}{1}}%
                        {\textcolor{nodeAcol!90!black}{A}}}
              \cdot
             \weights_{\textcolor{edge1col!90!black}{\!1}}
              \cdot
             \textcolor{nodeDcol!90!black}{D}\phantom{)}
        &)\\
    \text{R-GIN: }
      & \textcolor{nodeAcol!90!black}{A'} = 
        &\sigma&(
        &   \mathit{MLP}_{\textcolor{edge0col!90!black}{\!\circlearrowleft}} (\textcolor{nodeAcol!90!black}{A})
        &+& \mathit{MLP}_{\textcolor{edge1col!90!black}{\!1}}                (\textcolor{nodeBcol!90!black}{B})
        &+& \mathit{MLP}_{\textcolor{edge2col!90!black}{\!2}}                (\textcolor{nodeCcol!90!black}{C})
        &+& \mathit{MLP}_{\textcolor{edge1col!90!black}{\!1}}                (\textcolor{nodeDcol!90!black}{D})
        &)\\
    \text{GNN-MLP: }
    & \textcolor{nodeAcol!90!black}{A'} = 
      &\sigma&(
      &   \mathit{MLP}_{\textcolor{edge0col!90!black}{\!\circlearrowleft}} (\textcolor{nodeAcol!90!black}{A}\Vert\textcolor{nodeAcol!90!black}{A})
      &+& \mathit{MLP}_{\textcolor{edge1col!90!black}{\!1}}                (\textcolor{nodeBcol!90!black}{B}\Vert\textcolor{nodeAcol!90!black}{A})
      &+& \mathit{MLP}_{\textcolor{edge2col!90!black}{\!2}}                (\textcolor{nodeCcol!90!black}{C}\Vert\textcolor{nodeAcol!90!black}{A})
      &+& \mathit{MLP}_{\textcolor{edge1col!90!black}{\!1}}                (\textcolor{nodeDcol!90!black}{D}\Vert\textcolor{nodeAcol!90!black}{A})
      &)\\
    \text{RGDCN: }
      & \textcolor{nodeAcol!90!black}{A'} = 
        &\sigma&(
        &   \weights_{\textcolor{edge0col!90!black}{\circlearrowleft}, \textcolor{nodeAcol!90!black}{A}}
             \cdot
            \textcolor{nodeAcol!90!black}{A}\phantom{)}
        &+& \weights_{\textcolor{edge1col!90!black}{1}, \textcolor{nodeAcol!90!black}{A}}
              \cdot
            \textcolor{nodeBcol!90!black}{B}\phantom{)}
        &+& \weights_{\textcolor{edge2col!90!black}{2}, \textcolor{nodeAcol!90!black}{A}}
              \cdot
            \textcolor{nodeCcol!90!black}{C}\phantom{)}
        &+& \weights_{\textcolor{edge1col!90!black}{1}, \textcolor{nodeAcol!90!black}{A}}
              \cdot
            \textcolor{nodeDcol!90!black}{D}\phantom{)}
      &)\\
    \text{GNN-FiLM: }
      & \textcolor{nodeAcol!90!black}{A'} = 
        &\sigma&(
        &   \boldsymbol\beta_{\textcolor{edge0col!90!black}{\circlearrowleft}, \textcolor{nodeAcol!90!black}{A}}
             +
            \boldsymbol\gamma_{\textcolor{edge0col!90!black}{\circlearrowleft}, \textcolor{nodeAcol!90!black}{A}}
             \odot
            \weights_{\textcolor{edge0col!90!black}{\!\circlearrowleft}}
             \cdot
            \textcolor{nodeAcol!90!black}{A}\phantom{)}
        &+& \boldsymbol\beta_{\textcolor{edge1col!90!black}{1}, \textcolor{nodeAcol!90!black}{A}}
             +
            \boldsymbol\gamma_{\textcolor{edge1col!90!black}{1}, \textcolor{nodeAcol!90!black}{A}}
             \odot
            \weights_{\textcolor{edge1col!90!black}{\!1}} 
             \cdot
            \textcolor{nodeBcol!90!black}{B}\phantom{)}
        &+& \boldsymbol\beta_{\textcolor{edge2col!90!black}{2}, \textcolor{nodeAcol!90!black}{A}}
             +
            \boldsymbol\gamma_{\textcolor{edge2col!90!black}{2}, \textcolor{nodeAcol!90!black}{A}}
             \odot
            \weights_{\textcolor{edge2col!90!black}{\!2}} 
             \cdot
            \textcolor{nodeCcol!90!black}{C}\phantom{)}
        &+& \boldsymbol\beta_{\textcolor{edge1col!90!black}{1}, \textcolor{nodeAcol!90!black}{A}}
             +
            \boldsymbol\gamma_{\textcolor{edge1col!90!black}{1}, \textcolor{nodeAcol!90!black}{A}}
             \odot
            \weights_{\textcolor{edge1col!90!black}{\!1}} 
             \cdot
            \textcolor{nodeDcol!90!black}{D}\phantom{)}
        &)
   \end{alignat*}
  \end{minipage}}
  \caption{Computation of $\textcolor{nodeAcol!90!black}{A'}$ from \rF{fig:gnn-comp-sketch} in different GNN implementations
   (see main text for definitions of $\boldsymbol\alpha_v, \boldsymbol\beta_{\ell,v}, \boldsymbol\gamma_{\ell,v}, \weights_\ell, \mathit{MLP}_\ell$).
   Colours chosen to match colours of elements in \rF{fig:gnn-comp-sketch}; more colours in a term indicate more interactions between different elements.
   \label{fig:gnn-comp-eqs}
  }
\end{figure*}

\paragraph{\underline{G}raph \underline{N}eural \underline{N}etworks with \underline{F}eature-w\underline{i}se \underline{L}inear \underline{M}odulation (GNN-FiLM)}
In \rEq{eq:rgdcn}, the message passing layer is a linear transformation
conditioned on the target node representation, focusing on separate
chunks of the node representation at a time.
In the extreme case in which the dimension of each chunk is 1, this
method coincides with the ideas of \citet{perez2017film}, who propose to
use layers of element-wise affine transformations to modulate feature
maps in the visual question answering setting; there, a natural language
question is the input used to compute the affine transformation applied
to the features extracted from a picture.

In the graph setting, we can use each node's representation as an input
that determines an element-wise affine transformation of incoming messages,
allowing the model to dynamically up-weight and down-weight features based
on the information present at the target node of an edge.
This yields the following update rule, using a learnable function $g$ to
compute the parameters of the affine transformation.
\begin{align}
  \boldsymbol\beta^{(t)}_{\ell, v}, \boldsymbol\gamma^{(t)}_{\ell, v} & = g(\state{v}{t} \; ; \; \parameters_{g, \ell}) \nonumber\\
  \state{v}{t+1} & = 
    \sigma\left(
      \sum_{\typededge{u}{\ell}{v} \in \edges}
      \boldsymbol\gamma^{(t)}_{\ell, v} \odot \weights_{\!\ell} \state{u}{t} + \boldsymbol\beta^{(t)}_{\ell, v}
    \right) \label{eq:gnn_film}
\end{align}
The learnable parameters of the model are both the hypernetwork parameters
$\parameters_{g, \ell}$ and the weights $\weights_{\!\ell}$.
In practice, implementing $g$ as a single linear layer works well.

In the case of using a single linear layer, the resulting message passing
function is bilinear in source and target node representation,
as the message computation is centred around 
 $(\weights_{\!g,\ell} \state{v}{t}) \odot (\weights_{\!\ell} \state{u}{t})$.
This is the core difference to the (additive) interaction of source and target
node representations in models that use
 $\weights_{\!\ell} (\state{u}{t} \Vert \state{v}{t})$.

A simple toy example may illustrate the usefulness of such a mechanism:
assuming a graph of disjoint sets of nodes $\nodes_A$ and $\nodes_B$ and edge
types $1$ and $2$, a task may involve counting the number of $1$-neighbours
of $\nodes_A$ nodes and of $2$-neighbours of $\nodes_B$ nodes.
By setting 
 $\gamma_{1, v_a} = 1$, $\gamma_{2, v_a} = 0$ for $v_a \in \nodes_A$
and
 $\gamma_{1, v_b} = 0$, $\gamma_{2, v_b} = 1$ for $v_b \in \nodes_B$,
GNN-FiLM can solve this in a single layer.
Simpler approaches can solve this by counting $1$-neighbours and $2$-neighbours
separately in one layer and then projecting to the correct counter in the next
layer, but require more feature dimensions and layers for this.
As this toy example illustrates, a core capability of GNN-FiLM is to learn
to ignore graph edges based on the representation of target nodes.

Note that feature-wise modulation can also be viewed of an extension of
the gating mechanism of GRU or LSTM cells used in GGNNs.
Concretely, the ``forgetting'' of memories in a GRU/LSTM is similar to 
down-weighting messages computed for the self-loop edges and the gating of
the cell input is similar to the modulation of other incoming messages.
However, GGNNs apply this gating to the sum of all incoming messages (cf.
Eq. \rEq{eq:ggnn}), whereas in GNN-FiLM the modulation additionally
depends on the edge type, allowing for a more fine-grained gating mechanism.

Finally, a small implementation bug brought focus to the fact that applying
the non-linearity $\sigma$ after summing up messages from neighbouring nodes
can make it harder to perform tasks such as counting the number of neighbours
with a certain feature.
In experiments, applying the non-linearity before aggregation as in the
following update rule improved performance.
\begin{align}
  \state{v}{t+1} & = 
    l\left(
      \sum_{\typededge{u}{\ell}{v} \in \edges}
        \sigma\left(\boldsymbol\gamma^{(t)}_{\ell, v} \odot \weights_{\!\ell} \state{u}{t} + \boldsymbol\beta^{(t)}_{\ell, v}
              \right)
      \; ; \; \parameters_{l} \right) \label{eq:gnn_film2}
\end{align}
However, this means that the magnitude of node representations is
now dependent on the degree of nodes in the considered graph.
This can sometimes lead to instability during training, which can in
turn be controlled by adding an additional layer $l$ after message passing,
which can be
 a simple bounded nonlinearity (e.g. $\mathit{tanh}$),
 a fully connected layer,
 layer normalisation~\citep{ba2016layer},
 or any combination of these.

The different GNN definitions are illustrated by example in
\rF{fig:gnn-comp-eqs}, which shows how to compute a new representation of
node $\textcolor{nodeAcol!90!black}{A'}$ from \rF{fig:gnn-comp-sketch} for
all presented variants.

%\section{Related Work}
%\label{sect:relwork}
%\input{text/relwork}

\section{Evaluation}
\label{sect:evaluation}
\subsection{GNN Benchmark Tasks}
Due to the versatile nature of the GNN modelling formalism, many fundamentally
different tasks are studied in the research area and it should be noted
that good results on one task often do not transfer over to other tasks.
This is due to the widely varying requirements of different tasks, as the
following summary of tasks from the literature should illustrate.
\begin{itemize}
 \item Cora/Citeseer/Pubmed~\citep{sen08collective}:
  Each task consists of a single
  graph of $\sim{}10\,000$ nodes corresponding to documents and
  undirected (sic!) edges corresponding to references.
  The sparse $\sim{}1\,000$ node features are a bag of words representation
  of the corresponding documents.
  The goal is to assign a subset of nodes to a small number of classes.
  State of the art performance on these tasks is achieved with just two
  message propagation steps along graph edges, indicating that the graph
  structure is used to only a little degree.
% \item DIEL~\cite{Bingetal-TODO}: A bipartite single graph of 
%  $\sim{}4\,000\,000$ nodes representing medical entities and texts.
%  The goal is to assign TODO
 \item PPI~\citep{zitnik17predicting}: A protein-protein interaction
  dataset consisting of 24 graphs of $\sim{} 2\,500$ nodes corresponding
  to different human tissues.
  Each node has 50 features selected by domain experts and the goal is
  node-level classification, where each node may belong to several of the
  121 classes.
  State of the art performance on this task requires three propagation steps.
 \item QM9 property prediction~\citep{ramakrishnan14quantum}:
  $\sim{}130\,000$ graphs of $\sim 18$ nodes represent molecules, where
  nodes are atoms and undirected, typed edges are bonds between these
  atoms, different edge types indicating single/double/etc. bonds.
  The goal is to regress from each graph to a number of quantum chemical
  properties.
  State of the art performance on these tasks requires at least four
  propagation steps.
 \item VarMisuse~\citep{allamanis2018learning}: $\sim{}235\,000$ graphs of
  $\sim{}2500$ nodes each represent program fragments, where nodes are
  tokens in the program text and different edge types represent the
  program's abstract syntax tree, data flow between variables, etc.
  The goal is to select one of a set of candidate nodes per graph.
  State of the art performance requires at least six propagation
  steps.
\end{itemize}
Hence, tasks differ in
 the complexity of edges (from undirected and untyped to directed and many-typed),
 the size of the considered graphs,
 the size of the dataset,
 the importance of node-level vs. graph-level representations,
 and the number of required propagation steps.

This article includes results on the PPI, QM9 and VarMisuse tasks.
Preliminary experiments on the citation network data showed results that
were at best comparable to the baseline methods, but changes of a random
seed led to substantial fluctuations (mirroring the problems with
evaluation on these tasks reported by \citet{shchur18pitfalls}).

\subsection{Implementation}
To allow for a wider comparison, the implementation of GNN-FiLM is
accompanied by implementations of a range of baseline methods.
These include
 GGNN~\citep{li2015gated} (see Eq. \rEq{eq:ggnn}),
 R-GCN~\citep{schlichtkrull2017modeling} (see Eq. \rEq{eq:rgcn}),
 R-GAT~\citep{velickovic2018graph} (see Eq. \rEq{eq:gat}),
and
 R-GIN~\citep{hamilton2017inductive} (see Eq. \rEq{eq:rgin})\footnote{Note that
  Eq. \rEq{eq:gat} and Eq. \rEq{eq:rgin} define generalisations to different
  edge types not present in the original papers.}.
Additionally, GNN-MLP0 is a variant of R-GCN using a single linear layer
to compute the edge message from both source and target state (i.e., Eq.
\rEq{eq:gnn_edge_mlp} instantiated with an ``MLP'' without hidden layers),
and GNN-MLP1 is the same with a single hidden layer.
The baseline methods were re-implemented in TensorFlow and individually
tested to reach performance equivalent to results reported in their
respective source papers.
All code for the implementation of these GNNs is released on
 \url{https://github.com/Microsoft/tf-gnn-samples},
together with implementations of all tasks and scripts necessary
to reproduce the results reported in this paper.
This includes the hyperparameter settings found by search, which
are stored in \texttt{tasks/default\_hypers/} and are selected by
default on the respective tasks.
The code is designed to facilitate testing new GNN types on existing
tasks and easily adding new tasks, allowing for rapid evaluation of
new architectures.

Early on in the experiments, it became clear that the RGDCN approach
(Eq. \rEq{eq:rgdcn}) as presented is infeasible.
It is extremely sensitive to the parameter initialisation and hence
changes to the random seed lead to wild swings in the target
metrics.
Hence, no experimental results are reported for it in the following.
It is nonetheless included in the article (and the implementation)
to show the thought process leading to GNN-FiLM, as well as to allow
other researchers to build upon this.
In the following, GNN-FiLM refers to the formulation of Eq. 
\rEq{eq:gnn_film2}, which performed better than the variant of Eq.
\rEq{eq:gnn_film} across all experiments.
Somewhat surprisingly, the same trick (of moving the non-linearity
before the message aggregation step) did not help the other GNN
types.
For all models, using different layer weights for different propagation
steps performed better than using the same layer weights across several
propagation steps.

In all experiments, early stopping was used, i.e., the models were trained
until the target metric did not improve anymore for some additional epochs
(25 for PPI and QM9, 5 for VarMisuse).
The reported results on the held-out test data are averaged across the
results of a number of training runs, each starting from different
random parameter initialisations.

\subsection{Experimental Results}

\subsubsection{Protein-Protein Interactions (PPI)}
%\paragraph{PPI}
The models are first evaluated on the node-level classification
PPI task~\citep{zitnik17predicting}, following the dataset split from
earlier papers.
Training hence used a set of 20 graphs and validation and test sets
of two separate graphs each.
The graphs use two edge types: the dataset-provided untyped edges as well as
a fresh ``self-loop'' edge type to allows nodes to keep state across
propagation steps.

Hyperparameters for all models were selected based on results
from earlier papers and a small grid search of a number of author-selected
hyperparameter ranges (see \rApp{app:hypers} for details).
This resulted in three (R-GAT), four (GGNN, GNN-FiLM, GNN-MLP1, R-GCN), or five
(GNN-MLP0, R-GIN) layers (propagation steps) and a node representation size of 256
(GNN-MLP0, R-GIN) or 320 (all others).
All models use dropout on the node representations before all GNN layers,
with a keep ratio of 0.9.
After selecting hyperparameters, all models were trained ten
times with different random seeds on a NVidia V100.

\begin{table}
 \caption{\label{tab:ppi}GNN results on PPI task. GAT$^*$ result taken from \citet{velickovic2018graph}.}
 \vspace{-3ex}
 \begin{center}
 \begin{tabular}{@{}lS[table-format=1.3]@{}lr@{}}
  \toprule
   Model    & \multicolumn{2}{c}{Avg. Micro-F1} & Time (s) \\
  \midrule
   GAT$^*$  &           0.973 & $\pm 0.002$     & n/a   \\
   GGNN     &           0.990 & $\pm 0.001$     & 432.6 \\
   R-GCN    &           0.989 & $\pm 0.000$     & 759.0 \\
   R-GAT    &           0.989 & $\pm 0.001$     & 782.3 \\
   R-GIN    &           0.991 & $\pm 0.001$     & 704.8 \\
   GNN-MLP0 & \bfseries 0.992 & $\pm 0.000$     & 556.9 \\
   GNN-MLP1 & \bfseries 0.992 & $\pm 0.001$     & 479.2 \\
   GNN-FiLM & \bfseries 0.992 & $\pm 0.000$     & 308.1 \\
  \bottomrule
 \end{tabular}
 \end{center}
\end{table}
\rTab{tab:ppi} shows the micro-averaged F1 score on the
classification task on the test graphs, with standard deviations
and training times in seconds computed over the ten runs.
The results for all re-implemented models are better than
the results reported by \citet{velickovic2018graph} for the GAT model
(without edge types).
A cursory exploration of the reasons yielded three factors.
First, the generalisation to different edge types (cf. Eq. \rEq{eq:gat}) and the 
subsequent use of a special self-loop edge type helps R-GAT (and all
other models) significantly.
Second, using dropout between layers significantly improved the results.
Third, the larger node representation sizes (compared to 256 used by
\citet{velickovic2018graph}) improved the results again.
\rSC{app:ppi-gat-ablations} in the appendix shows this
in detail with ablation experiments.
Overall, the new GNN-FiLM improves slightly over the four baselines from the
literature, while converging substantially faster than all baselines, mainly
because it converges in significantly fewer training steps (approx.
150 epochs compared to 400-700 epochs for the other models).
Training curves for all models are included in \rSC{app:training-curves}
in the appendix.

\begin{table*}[t]
  \caption{\label{tab:qm9}GNN average error rates and standard deviations on QM9 target values.}
  \vspace{-1ex}  
  \resizebox{\textwidth}{!}{
  \begin{tabular}{@{}l@{\quad}S[table-format=2.2]@{ }lS[table-format=2.2]@{ }lS[table-format=2.2]@{ }lS[table-format=2.2]@{ }lS[table-format=2.2]@{ }lS[table-format=2.2]@{ }lS[table-format=2.2]@{ }l@{}}
   \toprule
    Property  & \multicolumn{2}{c}{GGNN} %\citep{li2015gated}}
              & \multicolumn{2}{c}{R-GCN} %\citep{schlichtkrull2017modeling}}
              & \multicolumn{2}{c}{R-GAT} %\citep{velickovic2018graph}}
              & \multicolumn{2}{c}{R-GIN} %\citep{xu2019how}}
              & \multicolumn{2}{c}{GNN-MLP0}
              & \multicolumn{2}{c}{GNN-MLP1}
              & \multicolumn{2}{c}{GNN-FiLM}\\
   \midrule
      mu &   3.85 & $\pm 0.16$  &   3.21 & $\pm 0.06$ &   2.68 & $\pm  0.06$ &   2.64 & $\pm  0.11$ &\bfseries  2.36 & $\pm 0.04$ &  2.44 & $\pm 0.12$ &  2.38 & $\pm 0.13$\\
   alpha &   5.22 & $\pm 0.86$  &   4.22 & $\pm 0.45$ &   4.65 & $\pm  0.44$ &   4.67 & $\pm  0.52$ &  4.27 & $\pm 0.36$ &  4.63 & $\pm 0.54$ &\bfseries  3.75 & $\pm 0.11$\\
    HOMO &   1.67 & $\pm 0.07$  &   1.45 & $\pm 0.01$ &   1.48 & $\pm  0.03$ &   1.42 & $\pm  0.01$ &  1.25 & $\pm 0.04$ &  1.29 & $\pm 0.06$ &\bfseries  1.22 & $\pm 0.07$\\
    LUMO &   1.74 & $\pm 0.06$  &   1.62 & $\pm 0.04$ &   1.53 & $\pm  0.07$ &   1.50 & $\pm  0.09$ &  1.35 & $\pm 0.04$ &  1.50 & $\pm 0.19$ &\bfseries  1.30 & $\pm 0.05$\\
     gap &   2.60 & $\pm 0.06$  &   2.42 & $\pm 0.14$ &   2.31 & $\pm  0.06$ &   2.27 & $\pm  0.09$ &  2.04 & $\pm 0.05$ &  2.06 & $\pm 0.10$ &\bfseries  1.96 & $\pm 0.06$\\
      R2 &  35.94 & $\pm 35.68$ &  16.38 & $\pm 0.49$ &  52.39 & $\pm 42.58$ &  15.63 & $\pm  1.40$ &\bfseries 14.86 & $\pm 1.62$ & 15.81 & $\pm 1.42$ & 15.59 & $\pm 1.38$\\
    ZPVE &  17.84 & $\pm 3.61$  &  17.40 & $\pm 3.56$ &  14.87 & $\pm  2.88$ &  12.93 & $\pm  1.81$ & 12.00 & $\pm 1.66$ & 14.12 & $\pm 1.10$ &\bfseries 11.00 & $\pm 0.74$\\
      U0 &   8.65 & $\pm 2.46$  &   7.82 & $\pm 0.80$ &   7.61 & $\pm  0.46$ &   5.88 & $\pm  1.01$ &  5.55 & $\pm 0.38$ &  6.94 & $\pm 0.64$ &\bfseries  5.43 & $\pm 0.96$\\
       U &   9.24 & $\pm 2.26$  &   8.24 & $\pm 1.25$ &   6.86 & $\pm  0.53$ &  18.71 & $\pm 23.36$ &  6.20 & $\pm 0.88$ &  7.00 & $\pm 1.06$ &\bfseries  5.95 & $\pm 0.46$\\
       H &   9.35 & $\pm 0.96$  &   9.05 & $\pm 1.21$ &   7.64 & $\pm  0.92$ &   5.62 & $\pm  0.81$ &  5.96 & $\pm 0.45$ &  7.98 & $\pm 0.88$ &\bfseries  5.59 & $\pm 0.57$\\
       G &   7.14 & $\pm 1.15$  &   7.00 & $\pm 1.51$ &   6.54 & $\pm  0.36$ &   5.38 & $\pm  0.75$ &\bfseries  5.09 & $\pm 0.57$ &  7.14 & $\pm 0.51$ &  5.17 & $\pm 1.13$\\
      Cv &   8.86 & $\pm 9.07$  &   3.93 & $\pm 0.48$ &   4.11 & $\pm  0.27$ &   3.53 & $\pm  0.37$ &\bfseries  3.38 & $\pm 0.20$ &  4.60 & $\pm 0.74$ &  3.46 & $\pm 0.21$\\
   Omega &   1.57 & $\pm 0.53$  &   1.02 & $\pm 0.05$ &   1.48 & $\pm  0.87$ &   1.05 & $\pm  0.11$ &\bfseries  0.84 & $\pm 0.02$ &  5.60 & $\pm 8.82$ &  0.98 & $\pm 0.06$\\
   \bottomrule
  \end{tabular}
  }
\end{table*}

\subsubsection{Quantum Chemistry (QM9)}
%\paragraph{QM9}
All models were additionally evaluated on graph-level regression
tasks on the QM9 molecule data set~\citep{ramakrishnan14quantum}, considering
thirteen different quantum chemical properties.
The $\sim\!\!130k$ molecular graphs in the dataset were split into training,
validation and test data by randomly selecting 10\,000 graphs for the 
latter two sets.
Additionally, another data split without a test set was used for the
hyperparameter search (see below).
The graphs use five edge types: the dataset-provided typed edges (single, double,
triple and aromatic bonds between atoms) as well as a fresh ``self-loop'' edge
type that allows nodes to keep state across propagation steps.
The evaluation differs from the setting reported by \citet{gilmer2017neural},
as no additional molecular information is encoded as edge features, nor are the
graphs augmented by master nodes or additional edges.\footnote{Adding 
  these features is straightforward, but orthogonal to the comparison
  of different GNN variants.}

Hyperparameters for all models were found using a staged search process.
First, 500 hyperparameter configurations were sampled from an author-provided
search space (see \rApp{app:hypers} for details) and run on the first three
regression tasks.
The top three configurations for each of these three tasks were then run
on all thirteen tasks and the final configuration was chosen as the one with
the lowest average mean absolute error across all properties, as evaluated
on the validation data of that dataset split.
This process led to eight layers / propagation steps for all models but GGNN and R-GIN,
which showed best performance with six layers.
Furthermore, all models used residual connections connecting every
second layer and GGNN, R-GCN, GNN-FiLM and GNN-MLP0 additionally used layer
normalisation (as in Eq. \rEq{eq:gnn_film2}).

Each model was trained for each of the properties separately five
times using different random seeds on compute nodes with NVidia P100
cards.
The average results of the five runs are reported in \rTab{tab:qm9},
with their respective standard deviations.%
% and the average time in minutes that a training run needed.
\footnote{Note that training sometimes did not converge (as visible
 in the large standard deviation).
 Removing these outliers, GGNN achieved $18.11 (\pm 1.62)$ and R-GAT
 achieved $17.66 (\pm 1.23)$ on R2;
 R-GIN has an average error rate of $7.04 (\pm 1.41)$ on U, and
 GNN-MLP1's result on the Omega task is $1.19 (\pm 0.08)$.}
The results indicate that the new GNN-FiLM model outperforms the
standard baselines on all tasks and the usually not considered
GNN-MLP variants on the majority of tasks.

\subsubsection{Variable Usage in Programs (VarMisuse)}
%\paragraph{VarMisuse}
Finally, the models were evaluated on the VarMisuse
task of \citet{allamanis2018learning}.
This task requires to process a graph representing an abstraction of a
program fragment and then select one of a few candidate nodes (representing
program variables) based on the representation of another node (representing
the location to use a variable in).
The experiments are performed using the released split of the dataset,
which contains $\sim130k$ training graphs, $\sim20k$ validation graphs
and two test sets: \textsc{SeenProjTest}, which contains $\sim55k$
graphs extracted from open source projects that also contributed data
to the training and validation sets, and \textsc{UnseenProjTest},
which contains $\sim30k$ graphs extracted from completely unseen
projects.

\begin{table*}[t]
  \caption{\label{tab:varmisuse}Accuracy on VarMisuse task. GGNN$^*$ result taken
   from appendix of \citet{allamanis2018learning}.}
  \vspace{-2ex}
  \begin{center}
  \begin{tabular}{
    l
    S[table-format=2.1,table-space-text-pre=x]@{\hspace{2ex}}S[table-format=1.1,table-number-alignment=left]
    @{\hspace{5ex}}
    S[table-format=2.1,table-space-text-pre=xx]@{\hspace{2ex}}S[table-format=1.1,table-number-alignment=left]
    @{\hspace{5ex}}
    S[table-format=2.1,table-space-text-pre=xxx]@{\hspace{2ex}}S[table-format=1.1,table-number-alignment=left]
    @{\hspace{5ex}}
    S[table-format=2.1,table-space-text-pre=xxxxx]@{\hspace{2ex}}S[table-format=1.1,table-number-alignment=left]
    }
   \toprule
    Model    &  \multicolumn{2}{c}{\textsc{Train}} &  \multicolumn{2}{c}{\textsc{Valid}} &  \multicolumn{2}{c}{\textsc{SeenProjTest}} &  \multicolumn{2}{c}{\textsc{UnseenProjTest}}\\
   \midrule
    GGNN$^*$ &                     n/a  &          &                     n/a  &          &                            84.0 & n/a &                                   74.1 & n/a \\
    GGNN     &                     87.5 &\pm 1.8\% &                     82.1 &\pm 0.9\% &                            85.7 &\pm 0.5\% &                              79.3 &\pm 1.2\% \\
    R-GCN    &                     88.7 &\pm 3.1\% &                     85.7 &\pm 1.6\% & \bfseries                  87.2 &\pm 1.5\% & \bfseries                    81.4 &\pm 2.3\% \\
    R-GAT    &                     90.4 &\pm 3.9\% &                     84.2 &\pm 1.0\% &                            86.9 &\pm 0.7\% &                              81.2 &\pm 0.9\% \\
    R-GIN    &                     93.4 &\pm 1.8\% &                     84.2 &\pm 1.0\% &                            87.1 &\pm 0.1\% &                              81.1 &\pm 0.9\% \\
    GNN-MLP0 &                     95.3 &\pm 2.4\% &                     83.4 &\pm 0.3\% &                            86.5 &\pm 0.2\% &                              80.5 &\pm 1.4\% \\
    GNN-MLP1 &                     94.7 &\pm 1.2\% &                     84.4 &\pm 0.4\% &                            86.9 &\pm 0.3\% & \bfseries                    81.4 &\pm 0.7\% \\
    GNN-FiLM &                     94.3 &\pm 1.0\% &                     84.6 &\pm 0.6\% &                            87.0 &\pm 0.2\% &                              81.3 &\pm 0.9\% \\
   \bottomrule
  \end{tabular}
  \end{center}
\end{table*}

Due to the inherent cost of training models on this dataset (\citet{balog19fast}
provide an in-depth performance analysis), a limited hyperparameter grid
search was performed, with only $\sim30$ candidate configurations for each
model (see \rApp{app:hypers} for details).
For each model, the configuration yielding the best results on the validation
data set fold was selected.
This led to six layers for GGNN and R-GIN, eight layers for R-GAT and GNN-MLP0, and ten
layers for the remaining models. Graph node hidden sizes were 128 for all
models but GGNN and R-GAT, which performed better with 96 dimensions.

The results, shown in \rTab{tab:varmisuse}, are somewhat surprising, as they
indicate a different ranking of model architectures as the results on PPI and
QM9, with R-GCN performing best.
All re-implemented baselines beat the results reported by \citet{allamanis2018learning},
who also reported that R-GCN and GGNN show very similar performance.
This is in spite of a simpler implementation of the task than in the original
paper, as it only uses the string labels of nodes for the representation and does
not use the additional type information provided in the dataset.
However, the re-implementation of the task uses the insights from \citet{cvitkovic2019open},
who use character CNNs to encode node labels and furthermore introduce extra
nodes for subtokens appearing in labels of different nodes, connecting them to
their sources (e.g., nodes labelled \code{openWullfrax} and \code{closeWullfrax}
are both connected to a fresh \code{Wullfrax} node).

A deeper investigation results showed that the more complex models seem
to suffer from significant overfitting to the training data, as can be seen in the
results for training and validation accuracy reported in \rTab{tab:varmisuse}.
A brief exploration of more aggressive regularisation methods (more dropout,
weight decay) showed no improvement.

A possible explanation would be that there are classes of examples that can
not be solve by one model architecture due to limits in its expressivity, but
that are solvable by other architectures.
To this end, we can consider the number of examples that can only be solved
by one model architecture, but not by another.
For example, an in-depth analysis shows that 2.2\% of the examples in
\textsc{SeenProjTest} are predicted correctly by (at least) one of the five
trained R-GCN models, whereas 2.7\% of examples could be predicted correctly
by one of the trained R-GIN models.
This indicates that R-GCN's better (average) results are not due to a
more expressive architecture, but that training is just slightly more
successful at finding parameters that work well across all examples.

Finally, the large variance in results on the validation set (especially
for R-GCN) makes it likely that the hyperparameter grid search with only one
training run per configuration did not yield the best configuration for each
model.

\section{Discussion \& Conclusions}
\label{sect:discussion}
After a review of existing graph neural network architectures, the idea of using
hypernetwork-inspired models in the graph setting was explored.
This led to two models: Graph Dynamic Convolutional Networks and GNNs with 
feature-wise linear modulation.
While RGDCNs seem to be impractical to train, experiments show that GNN-FiLM
outperforms the established baseline models from the literature.
However, extensive experiments have shown that the same holds for the simple
GNN-MLP definition, which is usally not considered in GNN evaluations.

The extensive experiments also show that a number of results from the
literature could benefit from more substantial hyperparameter search and are
often missing comparisons to a number of obvious baselines:
\begin{itemize}
    \item The results in \rTab{tab:ppi} indicate that GATs have no advantage
     over GGNNs or R-GCNs on the PPI task, which does not match the findings
     by \citet{velickovic2018graph}.
    \item The results in \rTab{tab:varmisuse} indicate that R-GCNs are 
     outperforming GGNNs substantially on the VarMisuse task, contradicting
     the findings of \citet{allamanis2018learning}.
    \item The GNN-MLP models are obvious implementations of the core GNN
     message passing principle that are often alluded to, but are not part
     of the usually considered set of baseline models.
     Nonetheless, experiments across all three tasks have shown that these
     methods outperform better-published techniques such as GGNNs, R-GCNs
     and GATs, without a substantial runtime penalty.
\end{itemize}
These results indicate that there is substantial value in independent
reproducibility efforts and comparisons that include ``obvious'' baselines,
matching the experiences from other areas of machine learning as well as
earlier work by \citet{shchur18pitfalls} on reproducing experimental results
for GNNs on citation network tasks.

%% Removed for preprint
\subsubsection*{Acknowledgments}
 The author wants to thank 
 Miltos Allamanis for the many discussions about GNNs and feedback on a draft of this article,
 Daniel Tarlow for helpful discussions and pointing to the FiLM idea,
 Pashmina Cameron for feedback on the implementation and this article,
 and
 Uri Alon and the anonymous reviewers for suggesting improvements to the paper.

\bibliography{bibliography}

\begin{thebibliography}{21}
\providecommand{\natexlab}[1]{#1}
\providecommand{\url}[1]{\texttt{#1}}
\expandafter\ifx\csname urlstyle\endcsname\relax
  \providecommand{\doi}[1]{doi: #1}\else
  \providecommand{\doi}{doi: \begingroup \urlstyle{rm}\Url}\fi

\bibitem[Allamanis et~al.(2018)Allamanis, Brockschmidt, and
  Khademi]{allamanis2018learning}
Allamanis, M., Brockschmidt, M., and Khademi, M.
\newblock Learning to represent programs with graphs.
\newblock In \emph{International Conference on Learning Representations
  (ICLR)}, 2018.

\bibitem[Ba et~al.(2016)Ba, Kiros, and Hinton]{ba2016layer}
Ba, L.~J., Kiros, R., and Hinton, G.~E.
\newblock Layer normalization.
\newblock \emph{CoRR}, abs/1607.06450, 2016.

\bibitem[Balog et~al.(2019)Balog, van Merri{\"{e}}nboer, Moitra, Li, and
  Tarlow]{balog19fast}
Balog, M., van Merri{\"{e}}nboer, B., Moitra, S., Li, Y., and Tarlow, D.
\newblock Fast training of sparse graph neural networks on dense hardware.
\newblock \emph{CoRR}, abs/1906.11786, 2019.

\bibitem[Busbridge et~al.(2019)Busbridge, Sherburn, Cavallo, and
  Hammerla]{busbridge19rgat}
Busbridge, D., Sherburn, D., Cavallo, P., and Hammerla, N.~Y.
\newblock Relational graph attention networks.
\newblock \emph{CoRR}, abs/1904.05811, 2019.

\bibitem[Cvitkovic et~al.(2019)Cvitkovic, Singh, and
  Anandkumar]{cvitkovic2019open}
Cvitkovic, M., Singh, B., and Anandkumar, A.
\newblock Open vocabulary learning on source code with a graph-structured
  cache.
\newblock In \emph{International Conference on Machine Learning (ICML)}, 2019.

\bibitem[Gilmer et~al.(2017)Gilmer, Schoenholz, Riley, Vinyals, and
  Dahl]{gilmer2017neural}
Gilmer, J., Schoenholz, S.~S., Riley, P.~F., Vinyals, O., and Dahl, G.~E.
\newblock Neural message passing for quantum chemistry.
\newblock In \emph{International Conference on Machine Learning (ICML)}, 2017.

\bibitem[Ha et~al.(2017)Ha, Dai, and Le]{ha2016HyperNetworks}
Ha, D., Dai, A.~M., and Le, Q.~V.
\newblock {HyperNetworks}.
\newblock In \emph{International Conference on Learning Representations
  (ICLR)}, 2017.

\bibitem[Hamilton et~al.(2017)Hamilton, Ying, and
  Leskovec]{hamilton2017inductive}
Hamilton, W.~L., Ying, R., and Leskovec, J.
\newblock Inductive representation learning on large graphs.
\newblock In \emph{Advances in Neural Information Processing Systems
  (NeurIPS)}, 2017.

\bibitem[Kipf \& Welling(2017)Kipf and Welling]{kipf2017semi}
Kipf, T.~N. and Welling, M.
\newblock Semi-supervised classification with graph convolutional networks.
\newblock In \emph{International Conference on Learning Representations}, 2017.

\bibitem[Li et~al.(2016)Li, Tarlow, Brockschmidt, and Zemel]{li2015gated}
Li, Y., Tarlow, D., Brockschmidt, M., and Zemel, R.
\newblock Gated graph sequence neural networks.
\newblock In \emph{International Conference on Learning Representations
  (ICLR)}, 2016.

\bibitem[Paliwal et~al.(2019)Paliwal, Loos, Rabe, Bansal, and
  Szegedy]{paliwal19graph}
Paliwal, A., Loos, S.~M., Rabe, M.~N., Bansal, K., and Szegedy, C.
\newblock Graph representations for higher-order logic and theorem proving.
\newblock \emph{CoRR}, abs/1905.10006, 2019.

\bibitem[Perez et~al.(2017)Perez, Strub, de~Vries, Dumoulin, and
  Courville]{perez2017film}
Perez, E., Strub, F., de~Vries, H., Dumoulin, V., and Courville, A.~C.
\newblock {FiLM}: Visual reasoning with a general conditioning layer.
\newblock In \emph{AAAI Conference on Artificial Intelligence}, 2017.

\bibitem[Ramakrishnan et~al.(2014)Ramakrishnan, Dral, Rupp, and
  Lilienfeld]{ramakrishnan14quantum}
Ramakrishnan, R., Dral, P.~O., Rupp, M., and Lilienfeld, O. A.~V.
\newblock Quantum chemistry structures and properties of 134 kilo molecules.
\newblock \emph{Scientific Data}, 1, 2014.

\bibitem[Schlichtkrull et~al.(2018)Schlichtkrull, Kipf, Bloem, van~den Berg,
  Titov, and Welling]{schlichtkrull2017modeling}
Schlichtkrull, M., Kipf, T.~N., Bloem, P., van~den Berg, R., Titov, I., and
  Welling, M.
\newblock Modeling relational data with graph convolutional network.
\newblock In \emph{Extended Semantic Web Conference (ESWC)}, 2018.

\bibitem[Selsam et~al.(2019)Selsam, Lamm, B{\"{u}}nz, Liang, de~Moura, and
  Dill]{selsam19learning}
Selsam, D., Lamm, M., B{\"{u}}nz, B., Liang, P., de~Moura, L., and Dill, D.~L.
\newblock Learning a {SAT} solver from single-bit supervision.
\newblock In \emph{International Conference on Learning Representations
  (ICLR)}, 2019.

\bibitem[Sen et~al.(2008)Sen, Namata, Bilgic, Getoor, Galligher, and
  Eliassi-Rad]{sen08collective}
Sen, P., Namata, G., Bilgic, M., Getoor, L., Galligher, B., and Eliassi-Rad, T.
\newblock Collective classification in network data.
\newblock \emph{AI magazine}, 29, 2008.

\bibitem[Shchur et~al.(2018)Shchur, Mumme, Bojchevski, and
  G{\"{u}}nnemann]{shchur18pitfalls}
Shchur, O., Mumme, M., Bojchevski, A., and G{\"{u}}nnemann, S.
\newblock Pitfalls of graph neural network evaluation.
\newblock \emph{CoRR}, abs/1811.05868, 2018.

\bibitem[Veli{\v{c}}kovi{\'{c}} et~al.(2018)Veli{\v{c}}kovi{\'{c}}, Cucurull,
  Casanova, Romero, Li{\`{o}}, and Bengio]{velickovic2018graph}
Veli{\v{c}}kovi{\'{c}}, P., Cucurull, G., Casanova, A., Romero, A., Li{\`{o}},
  P., and Bengio, Y.
\newblock {Graph Attention Networks}.
\newblock In \emph{International Conference on Learning Representations
  (ICLR)}, 2018.

\bibitem[Wu et~al.(2019)Wu, Fan, Baevski, Dauphin, and Auli]{wu2018pay}
Wu, F., Fan, A., Baevski, A., Dauphin, Y., and Auli, M.
\newblock Pay less attention with lightweight and dynamic convolutions.
\newblock In \emph{International Conference on Learning Representations
  (ICLR)}, 2019.

\bibitem[Xu et~al.(2019)Xu, Hu, Leskovec, and Jegelka]{xu2019how}
Xu, K., Hu, W., Leskovec, J., and Jegelka, S.
\newblock How powerful are graph neural networks?
\newblock In \emph{International Conference on Learning Representations
  (ICLR)}, 2019.

\bibitem[Zitnik \& Leskovec(2017)Zitnik and Leskovec]{zitnik17predicting}
Zitnik, M. and Leskovec, J.
\newblock Predicting multicellular function through multi-layer tissue
  networks.
\newblock \emph{Bioinformatics}, 33, 2017.

\end{thebibliography}
\bibliographystyle{icml2020}

\newpage
\appendix
\section{Hyperparameter search spaces}
\label{app:hypers}
\subsection{PPI}
For all models, a full grid search considering all combinations of the following parameters was performed:
\begin{itemize}
    \item $\code{hidden\_size} \in \{ 192, 256, 320 \}$ - size of per-node representations.
    \item $\code{graph\_num\_layers} \in \{ 2, 3, 4, 5\}$ - number of propagation steps / layers.
    \item $\code{graph\_layer\_input\_dropout\_keep\_prob} \in \{ 0.8, 0.9, 1.0 \}$ - dropout applied before propagation steps.
\end{itemize}

\subsection{QM9}
For all models, 500 configurations were considered, sampling hyperparameter settings uniformly from the following options:
\begin{itemize}
    \item $\code{hidden\_size} \in \{ 64, 96, 128 \}$ - size of per-node representations.
    \item $\code{graph\_num\_layers} \in \{ 4, 6, 8\}$ - number of propagation steps / layers.
    \item $\code{graph\_layer\_input\_dropout\_keep\_prob} \in \{ 0.8, 0.9, 1.0 \}$ - dropout applied before propagation steps.
    \item $\code{layer\_norm} \in \{ \mathit{True}, \mathop{False} \}$ - decided if layer norm is applied after each propagation step.
    \item $\code{dense\_layers} \in \{ 1, 2, 32 \}$ - insert a fully connected layer applied to node representations between every \code{dense\_layers} propagation steps. (32 effectively turns this off)
    \item $\code{res\_connection} \in \{ 1, 2, 32 \}$ - insert a residual connection between every \code{res\_connection} propagation steps. (32 effectively turns this off)
    \item $\code{graph\_activation\_function} \in \{ \mathit{relu}, \mathit{leaky\_relu}, \mathit{elu}, \mathit{gelu}, \mathit{tanh} \}$ - non-linearity applied after message passing.
    \item $\code{optimizer} \in \{ \mathit{RMSProp}, \mathit{Adam} \}$ - optimizer used (with TF 1.13.1 default parameters).
    \item $\code{lr} \in [0.0005, 0.001]$ - learning rate.
    \item $\code{cell} \in \{ \mathit{RNN}, \mathit{GRU}, \mathit{LSTM} \}$ - gated cell used for GGNN.
    \item $\code{num\_heads} \in \{ 4, 8, 16 \}$ - number of attention heads used for R-GAT.
\end{itemize}

\subsection{VarMisuse}
For all models, a full grid search considering all combinations of the following parameters was performed:
\begin{itemize}
    \item $\code{hidden\_size} \in \{ 64, 96, 128 \}$ - size of per-node representations.
    \item $\code{graph\_num\_layers} \in \{ 6, 8, 10 \}$ - number of propagation steps / layers.
    \item $\code{graph\_layer\_input\_dropout\_keep\_prob} \in \{ 0.8, 0.9, 1.0 \}$ - dropout applied before propagation steps.
    \item $\code{cell} \in \{ \mathit{GRU}, \mathit{LSTM} \}$ - gated cell used for GGNN.
    \item $\code{num\_heads} \in \{ 4, 8 \}$ - number of attention heads used for R-GAT.
\end{itemize}

\section{R-GAT Ablations on PPI}
\label{app:ppi-gat-ablations}

\begin{table}
  \caption{\label{tab:ppi-gat-ablations}(R-)GAT ablation results on PPI task.}
  \vspace{-3ex}
  \begin{center}
  \begin{tabular}{@{}lS[table-format=1.3]@{}l@{}}
   \toprule
    Model    & \multicolumn{2}{c}{Avg. Micro-F1}\\
   \midrule
    GAT (dropout$=0.0$, dim$=256$) & 0.924 & $\pm 0.004$ \\
    GAT (dropout$=0.2$, dim$=256$) & 0.928 & $\pm 0.005$ \\
    GAT (dropout$=0.0$, dim$=320$) & 0.942 & $\pm 0.004$ \\
    GAT (dropout$=0.2$, dim$=320$) & 0.953 & $\pm 0.002$ \\
    R-GAT (dropout$=0.0$, dim$=256$) & 0.986 & $\pm 0.001$ \\
    R-GAT (dropout$=0.2$, dim$=256$) & 0.988 & $\pm 0.001$ \\
    R-GAT (dropout$=0.0$, dim$=320$) & 0.988 & $\pm 0.001$ \\
    R-GAT (dropout$=0.2$, dim$=320$) & 0.989 & $\pm 0.001$ \\
   \bottomrule
  \end{tabular}
  \end{center}
\end{table}

As discussed in the main text, the R-GAT model implemented used in the
experiments of this paper significantly outperforms the GAT model from
\citet{velickovic2018graph}.
To understand the causes of this, a number of ablation experiments were
performed, whose results are shown in \rTab{tab:ppi-gat-ablations}.

Note that while the ``GAT (dropout$=0.0$, dim$=256$)'' configuration
roughly corresponds to the GAT ablations used by \citet{velickovic2018graph},
there are two major differences that explain the performance difference.
First, the model here uses a linear layer to map the input node features
into a representation that has the same size as the hidden layers of the
GNN, while the original GAT implementation directly uses the input
representation.
Second, the re-implementation uses a linear layer applied to the final
node representation to obtain logits, whereas the original GAT model uses
the final GNN layer to directly compute the logits from the messages.

In the ablation results, the biggest jump in performance is caused by
switching from the original GAT formulation to the R-GAT formulation
from \rEq{eq:gat}. In the case of the PPI task, this only distinguishes
edges present in the dataset from the newly-introduced self-loop
edges used to keep state at a node, which seems to be of great importance
for model performance.

\section{Training Curves}
\label{app:training-curves}

\begin{figure}[t]
  \includegraphics[width=\columnwidth]{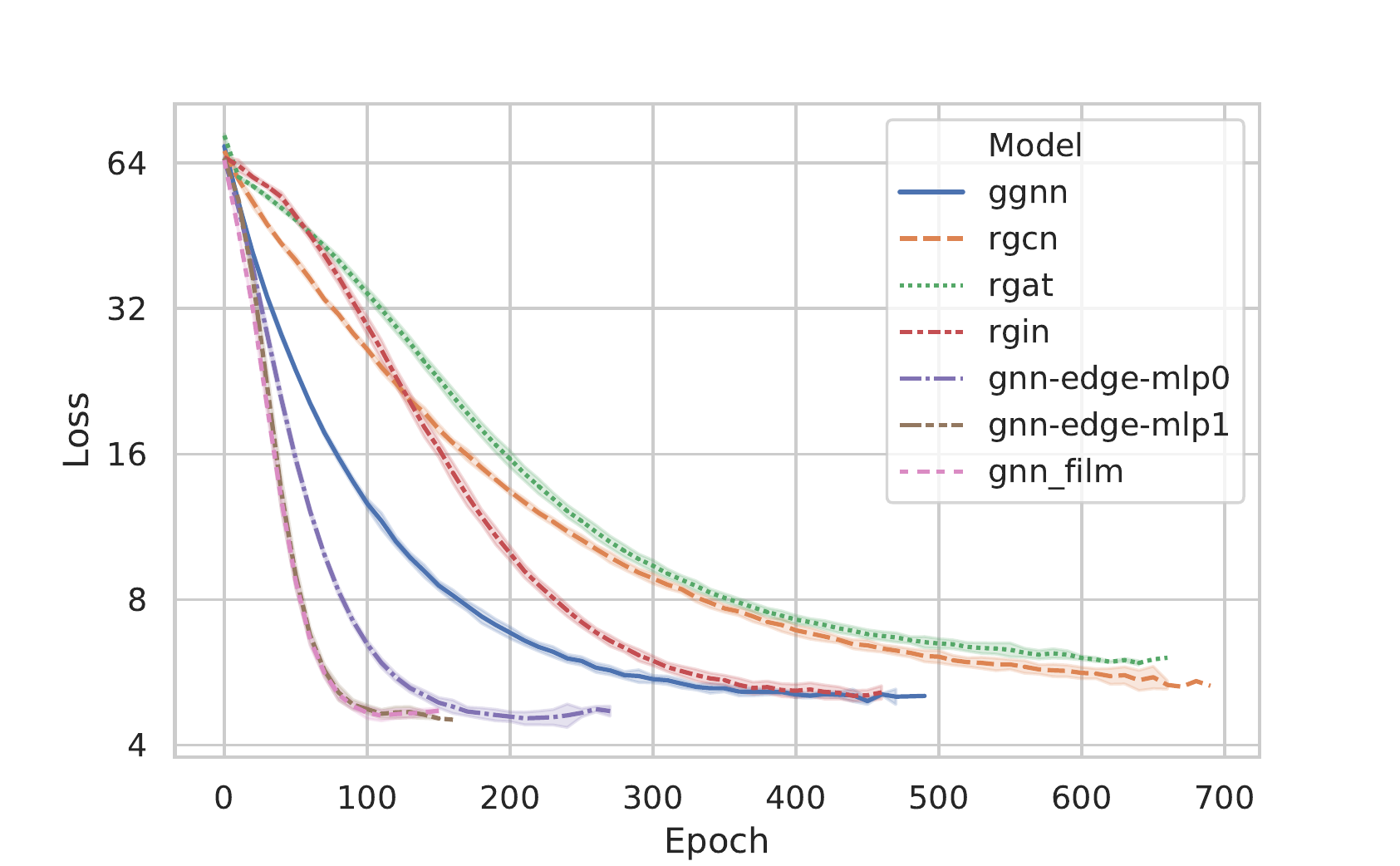}
  \caption{Validation loss of different models during training for PPI task.\label{fig:ppi-loss-curve}}
\end{figure}

The loss on the validation set during training of all considered models on the PPI task
is shown in \rF{fig:ppi-loss-curve}, with (barely visible) confidence intervals obtained
by using the runs for all 10 considered different random seeds.
The very fast convergence of the GNN-FiLM model becomes visible here.
The GNN-MLP0 model converges equally fast in terms of training steps, but requires more time per
epoch, and hence is slower in wall clock time, as seen in \rTab{tab:ppi}.

\begin{figure}[t]
  \includegraphics[width=\columnwidth]{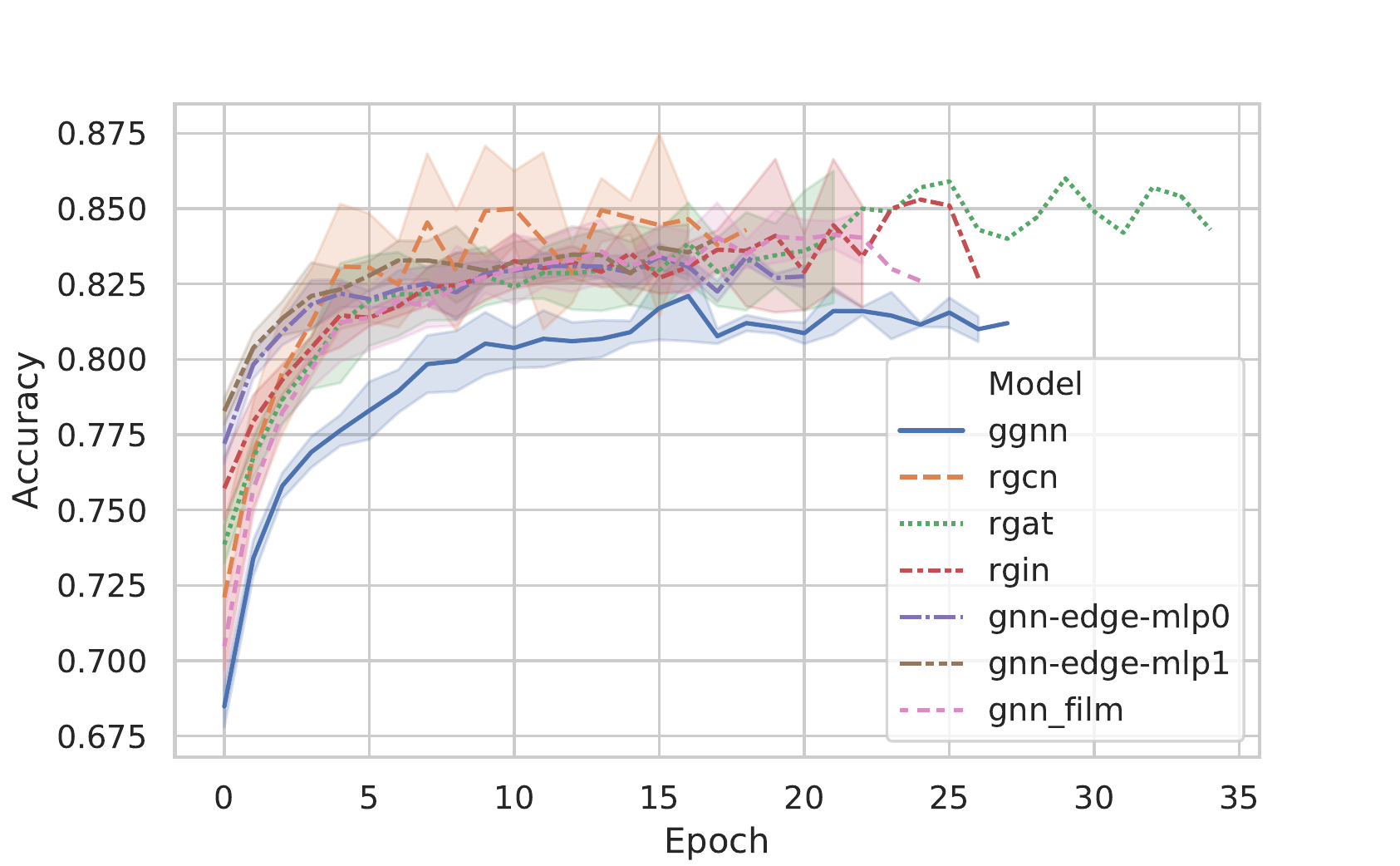}
  \caption{Accuracy on validation data of different models during training for VarMisuse task.\label{fig:varmisuse-acc-curve}}
\end{figure}

The accuracy on the validation set during training of all considered models on the VarMisuse
task is shown in \rF{fig:varmisuse-acc-curve}, with confidence intervals obtained by using
the runs for all 5 considered random seeds.

\end{document}